\documentclass[conference,letterpaper]{IEEEtran}
\IEEEoverridecommandlockouts
\usepackage{cite}
\usepackage{amsmath,amssymb}
\usepackage{graphicx}
\usepackage{booktabs,multirow,array}
\usepackage{xcolor}
\usepackage{tcolorbox}
\usepackage[hidelinks]{hyperref}

\title{Semantic-Guided Geometric Decoding for Token-Efficient Visual Grounding}
\author{
\IEEEauthorblockN{Zhaojie Chen,\quad Junyan Yang,\quad Le Yang\IEEEauthorrefmark{1},\quad Qi Dang,\quad Yucheng Huang\\
Qianyi Che,\quad Zhaoqiang Chen,\quad Xinqi Zhu,\quad Qiang Zhao}
\IEEEauthorblockA{Huawei Technologies Co., Ltd.}
\thanks{\IEEEauthorrefmark{1}Le Yang is the corresponding author. Email: \href{mailto:yangle47@huawei.com}{yangle47@huawei.com}.}
}
\hypersetup{
  pdftitle={Semantic-Guided Geometric Decoding for Token-Efficient Visual Grounding},
  pdfauthor={Zhaojie Chen, Junyan Yang, Le Yang, Qi Dang, Yucheng Huang, Qianyi Che, Zhaoqiang Chen, Xinqi Zhu, Qiang Zhao}
}

\begin{document}
\pagestyle{plain}
\maketitle
\thispagestyle{plain}
\begin{abstract}

Multimodal large language models (MLLMs) provide a unified interface for language-conditioned visual perception. In pixel-to-sequence grounding, autoregressive coordinate generation requires multiple sequential decoding steps per box, increasing inference cost with the number of instances. Pixel-to-embedding methods reduce this overhead but lack explicit spatial encoding when using a shared special token.
We introduce SemGeo, a token-efficient visual grounding framework that represents each instance with a single grid token for coarse localization. A geometric decoder reuses visual features and hidden states to recover precise discrete box coordinates in parallel. To train proposal generation and box localization together, we introduce Gaussian soft labels that encode coordinate distances during supervised fine-tuning (SFT). Reinforcement post-training further combines box-set rewards with coordinate supervision on model-generated proposals. This stage trains the language model and geometric decoder on their own predictions to mitigate the mismatch between teacher-forced training and inference.
Across eight benchmarks spanning object detection, referring object detection, and document layout grounding, SemGeo achieves the highest F1@mIou among compared MLLMs on six benchmarks. It also delivers $2.07\times$ and $1.25\times$ the mean box-generation throughput of representative grounding baselines under vLLM and Transformers, respectively, while achieving higher accuracy in both comparisons.

\end{abstract}

\section{Introduction}
\label{sec:intro}

\begin{figure}[!t]
\centering
\includegraphics[width=0.9\columnwidth]{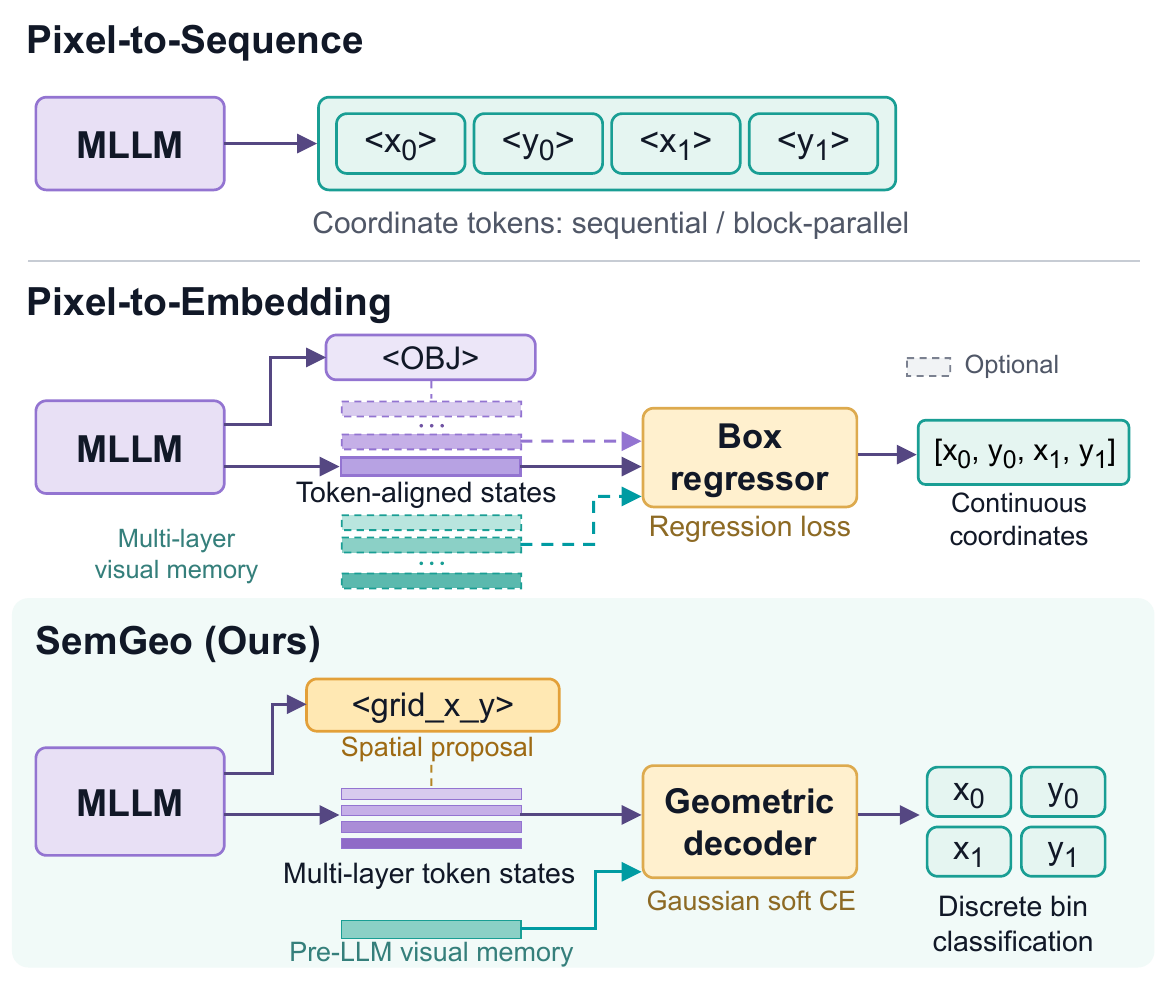}
\caption{\textbf{Comparison of representative grounding interfaces.} Top: coordinate-token prediction, sequential~\cite{rex_omni} or block-parallel~\cite{locate_anything}. Middle: token-triggered box regression~\cite{next_chat,perception_gpt}. Bottom: SemGeo uses grid tokens for coarse spatial supervision and predicts fine coordinates through parallel classification. \texttt{<OBJ>} denotes a generic localization token; dashed blocks indicate optional features.}
\label{fig:grounding_paradigms}
\end{figure}

Visual grounding connects language understanding with spatial prediction: a model must identify the entities requested by a user and localize their instances in an image. Detection has progressed from closed-set architectures~\cite{faster_rcnn,detr,dino} to language-conditioned open-set models~\cite{glip,grounding_dino,yolo_world}. Multimodal large language models (MLLMs) further extend this interface to natural-language instructions and referring expressions~\cite{shikra,kosmos_2,ferret,griffon,qwen3_vl}. Unified grounding systems such as Rex-Omni~\cite{rex_omni} and LocateAnything~\cite{locate_anything} demonstrate its applicability across object detection, referring object detection, and document layout analysis. These applications demand both accurate localization and efficient inference, particularly when detecting multiple instances in an image.

A critical design is how bounding boxes are represented and predicted, as illustrated in Fig.~\ref{fig:grounding_paradigms}. Following Pix2Seq's sequence-modeling formulation~\cite{pix2seq}, pixel-to-sequence methods represent boxes as textual or quantized coordinate tokens~\cite{shikra,kosmos_2,rex_omni}. Standard next-token prediction generates these tokens sequentially, while LocateAnything~\cite{locate_anything} reduces the decoding cost through box-aligned parallel token prediction. Pixel-to-embedding methods use contextualized representations for spatial decoding. NExT-Chat~\cite{next_chat} and PerceptionGPT~\cite{perception_gpt} predict boxes through dedicated regressors, while other approaches select from candidate boxes generated by dedicated object detectors~\cite{groma,chat_rex,vgent}.

Although existing methods have improved grounding accuracy and efficiency, pixel-to-sequence methods still represent each box with multiple output tokens~\cite{rex_omni,locate_anything}. Pixel-to-embedding methods can reduce the token overhead of coordinate generation, but a shared special token does not explicitly encode spatial location. InstructSAM~\cite{instruct_sam} identifies a related challenge in explicitly distinguishing instances during multi-instance segmentation. Beyond the output representation, learning precise localization also depends on the training objective and the context used during generation. One-hot coordinate targets do not encode distances between coordinate values~\cite{rex_omni}, while teacher-forced training uses ground-truth prefixes instead of the model's previous predictions, introducing exposure bias~\cite{mixer}.

To address these limitations, we introduce \textbf{SemGeo}, a token-efficient grounding framework that generates one grid token per instance for coarse localization and uses a geometric decoder to predict precise discrete box coordinates in parallel. As shown in Fig.~\ref{fig:grounding_paradigms} (bottom), SemGeo replaces the generic special token with a spatial grid token, encoding each instance's coarse location in its prediction target. To train proposal generation and box localization together, we develop a pipeline that incorporates coordinate distances into supervised fine-tuning (SFT). Reinforcement post-training then combines Group Relative Policy Optimization (GRPO)~\cite{deepseek_math} with coordinate supervision on model-generated proposals, adapting both the MLLM and geometric decoder to mitigate exposure bias. Experiments show that, under vLLM, SemGeo achieves higher accuracy than the Qwen3-VL-2B baseline with 87.6\% fewer output tokens and 85.4\% lower mean generation time.

Our main contributions are summarized as follows:
\begin{itemize}
	\item \textbf{Token-Efficient Coarse-to-Fine Grounding}: We introduce a coarse-to-fine grounding framework in which the language model generates one spatially supervised grid token per instance, while the geometric decoder predicts precise box coordinates in parallel. This division of labor reduces sequential backbone decoding and accelerates inference.
	
	\item \textbf{Training Pipeline for Coarse-to-Fine Grounding}: We develop a training pipeline that incorporates coordinate distances into supervised fine-tuning. Reinforcement post-training then adapts both the MLLM and geometric decoder to model-generated proposals to mitigate exposure bias.
	
	\item \textbf{Competitive Accuracy and Inference Speed}: Across eight widely used benchmarks, SemGeo exceeds LocateAnything by 1.70\%--4.70\% in F1@mIou on all six detection and layout benchmarks. It achieves $1.25\times$ the mean boxes per second (mean BPS) of LocateAnything under Transformers and $2.07\times$ that of Rex-Omni under vLLM, with higher accuracy in both paired comparisons.
\end{itemize}

\section{Related Work}
\label{sec:related_work}

Vision-language grounding methods differ in how they translate multimodal representations into spatial outputs. We review pixel-to-sequence and pixel-to-embedding interfaces, followed by parallel decoding strategies.

\paragraph{Pixel-to-Sequence Grounding.}
Pixel-to-sequence methods formulate localization as token generation. Pix2Seq~\cite{pix2seq} represents detections through discrete category and coordinate tokens, while SeqTR~\cite{seqtr} models referring boxes and masks as coordinate sequences. Shikra~\cite{shikra} and Kosmos-2~\cite{kosmos_2} extend spatial outputs to multimodal language interaction through textual coordinates and region-associated location tokens, respectively. Florence-2~\cite{florence_2} unifies perception tasks through sequence outputs, while Rex-Omni~\cite{rex_omni} uses dedicated quantized coordinate tokens for grounding. GETok~\cite{getok} represents box corners with grid tokens and refines their locations through autoregressively generated offset tokens, retaining token-based generation for fine localization. Under standard next-token prediction, these coordinate sequences require multiple sequential decoding steps per box, increasing inference latency in multi-instance scenes.

\paragraph{Pixel-to-Embedding Grounding.}
Pixel-to-embedding methods connect language understanding to spatial decoders through contextualized representations. NExT-Chat~\cite{next_chat} and PerceptionGPT~\cite{perception_gpt} decode embeddings into boxes or masks. PaDT~\cite{padt} groups image-derived visual reference tokens and decodes their last-layer representations into boxes and masks. VisionLLM v2~\cite{vision_llm_v2} uses learned super-link queries to condition task decoders, including Grounding DINO for detection, while VGent~\cite{vgent} uses layer-wise MLLM representations to select target boxes from proposals produced by off-the-shelf detectors through binary classification. LISA~\cite{lisa} and PixelLM~\cite{pixel_lm} decode segmentation-token states into masks. Qwen3-VL-Seg~\cite{qwen3_vl_seg} and WeVisOne-VL~\cite{wevisone_vl} further combine MLLM-predicted boxes, token representations, and multi-level visual features for box-guided mask decoding. 
These approaches obtain spatial information from contextualized representations, visual features, or explicit box priors. However, when the same special token is used across instances, the token itself does not explicitly encode spatial location. InstructSAM~\cite{instruct_sam} reports repeated or unstable masks when shared segmentation tokens lack explicit instance binding. These observations motivate incorporating spatial cues into token prediction for multi-instance grounding.

\begin{figure*}[!t]
	\centering
	\includegraphics[width=0.925\textwidth]{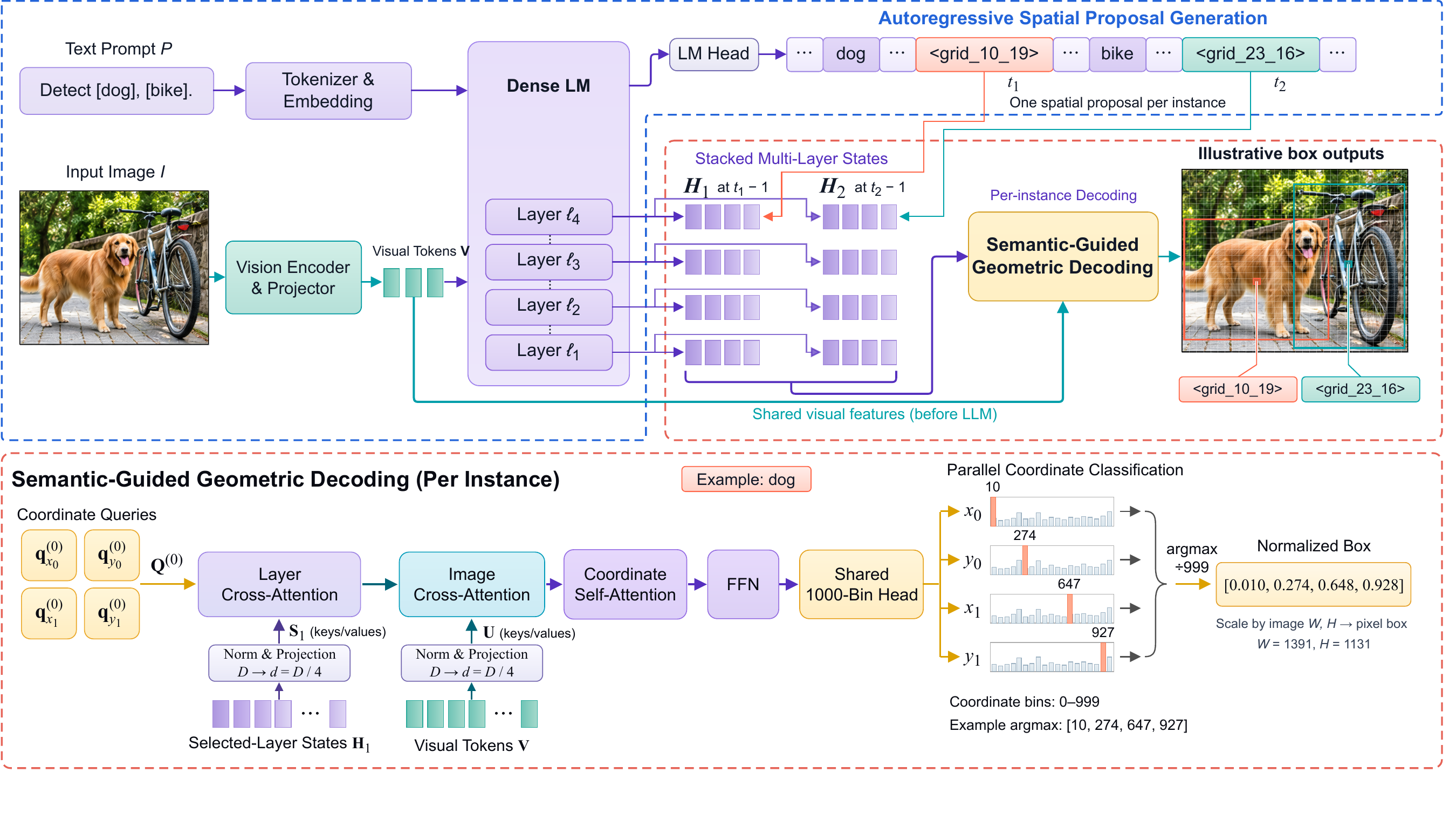}
    \caption{\textbf{Overview of SemGeo.} Blue and red dashed outlines denote Autoregressive Spatial Proposal Generation and Semantic-Guided Geometric Decoding, respectively. The dog and bike proposals at $t_1$ and $t_2$ use multi-layer states $\mathbf H_1$ and $\mathbf H_2$ from the preceding positions, together with shared visual features. Highlighted cells and boxes illustrate coarse proposals and decoded geometry. Bottom: Queries decode the dog box ($i=1$) via semantic and visual cross-attention, coordinate interaction, and parallel classification.}
	\label{fig:model_framework}
\end{figure*}

\paragraph{Blockwise Parallel and Speculative Decoding.}
Blockwise parallel decoding predicts multiple future tokens in parallel and accepts the prefix verified by an autoregressive model~\cite{blockwise_parallel_decoding}. Speculative decoding similarly separates drafting from target-model verification~\cite{speculative_decoding}. Multi-token prediction can supply candidate tokens through additional prediction heads or modules~\cite{multi_token_prediction,deepseek_v3}; Medusa~\cite{medusa} combines auxiliary heads with tree-attention verification.
In visual grounding, LocateAnything~\cite{locate_anything} predicts box-aligned blocks with block-causal attention and uses format and spatial-confidence checks to trigger autoregressive re-decoding in Hybrid mode.
LOCUS~\cite{locus} drafts a box with a regression head. The same backbone then verifies the corresponding coordinate tokens in parallel and iteratively refines the draft. These strategies reduce sequential decoding steps, but their speedup depends on draft quality and the computation required for verification or re-decoding.

\section{Method}
\label{sec:method}

\subsection{Framework Overview}
\label{sec:framework_overview}

Given an image $I$ and a prompt $P$, SemGeo predicts a set $\hat{\mathcal Y}=\{(s_i,\hat{\mathbf b}_i)\}_{i=1}^{N}$, where $N$ is the number of predicted instances and $i$ indexes them. For instance $i$, $s_i$ is a category or referring expression, and $\hat{\mathbf b}_i\in[0,1]^4$ is its normalized bounding box.
As shown in Fig.~\ref{fig:model_framework}, our framework comprises Autoregressive Spatial Proposal Generation (blue dashed outlines) and Semantic-Guided Geometric Decoding (red dashed outlines), implemented by an MLLM with parameters $\theta$ and a geometric decoder with parameters $\phi$, respectively.

\subsection{Autoregressive Spatial Proposal Generation}
\label{sec:spatial_proposals}

Autoregressive coordinate decoding requires the MLLM to enumerate instances while sequentially predicting their precise coordinates~\cite{rex_omni}. We move fine-grained coordinate prediction out of this autoregressive stream and represent each instance with a single spatial proposal token. Each \texttt{<grid\_x\_y>} indicates the instance's box-center cell on a $G_x\times G_y$ image grid, providing coarse spatial supervision. Conditioned on the image and prompt, the MLLM generates category names or referring expressions, delimiters, and proposal tokens through standard next-token prediction. It determines which instances to return and when to stop.
An illustrative response for the dog and bike in Fig.~\ref{fig:model_framework} is shown below:
\begingroup
\definecolor{semgeoExampleInput}{HTML}{DDEBFA}
\definecolor{semgeoExampleOutput}{HTML}{F7F8FA}
\definecolor{semgeoExampleFrame}{HTML}{8DA3BE}
\definecolor{semgeoExampleGrid}{HTML}{FFD2C7}
\definecolor{semgeoExampleGridBike}{HTML}{B5E0D8}
\begin{tcolorbox}[
    width=\linewidth,
    colback=semgeoExampleOutput,
    colframe=semgeoExampleFrame,
    colbacktitle=semgeoExampleInput,
    coltitle=black,
    fonttitle=\small,
    fontupper=\small,
    title={\textbf{Input:}\enspace\texttt{Detect [dog], [bike].}},
    boxrule=0.4pt,
    titlerule=0.4pt,
    arc=1mm,
    boxsep=0pt,
    left=4pt,right=4pt,top=4pt,bottom=4pt,
    toptitle=3pt,bottomtitle=3pt,
    before skip=4pt,after skip=4pt
]
\textbf{Output} (autoregressive text)\par\smallskip
\begingroup
\ttfamily
\setlength{\fboxsep}{1pt}
\noindent<|object\_ref\_start|>dog<|object\_ref\_end|>\\[2pt]
<|box\_start|>\colorbox{semgeoExampleGrid}{<grid\_10\_19>}<|box\_end|>,\\[2pt]
<|object\_ref\_start|>bike<|object\_ref\_end|>\\[2pt]
<|box\_start|>\colorbox{semgeoExampleGridBike}{<grid\_23\_16>}<|box\_end|>
\par\endgroup
\end{tcolorbox}
\endgroup
Instances in the same grid cell are represented by separate occurrences of the same grid token.

\subsection{Semantic-Guided Geometric Decoding}
\label{sec:geometric_decoder}

A coarse proposal does not specify the object's extent. To recover fine geometry for each instance, four coordinate-specific queries combine the hidden states used to predict its proposal token with image evidence in a reduced-dimensional space.

\paragraph{Feature extraction and projection.}
We select four evenly spaced language layers, including the final layer, to combine information from different depths. Let $t_i$ denote the sequence position of the proposal token for instance $i$. We collect hidden states at position $t_i-1$, which is used to predict that token. Let $\mathcal L$ denote the selected layer indices and $L=4$ their number. Stacking these states gives the feature matrix for instance $i$,
$\mathbf H_i=\operatorname{Stack}_{\ell\in\mathcal L}[\mathbf h_{t_i-1}^{(\ell)}]\in\mathbb R^{L\times D}$, whose rows contain the $D$-dimensional hidden states from the selected layers.

Visual features $\mathbf V\in\mathbb R^{N_v\times D_v}$ are taken after the visual projector, before the language backbone, where $N_v$ is the number of visual tokens and $D_v$ is their feature dimension. We normalize and project the semantic and visual features:
\begin{equation}
\begin{aligned}
\mathbf S_i &= \operatorname{Stack}_{\ell\in\mathcal L}
 [f_\ell(\operatorname{RN}_\ell(\mathbf h_{t_i-1}^{(\ell)}))],\\
\mathbf U &= f_v(\operatorname{RN}_v(\mathbf V)).
\end{aligned}
\label{eq:projected_memories}
\end{equation}
Here $\operatorname{RN}$ denotes RMS normalization; $\mathbf S_i\in\mathbb R^{L\times d}$ and $\mathbf U\in\mathbb R^{N_v\times d}$ share a reduced width $d$. In our backbone, $D_v=D$, and we set $d=D/4$ to reduce both feature sets' channel dimensions fourfold while retaining all visual tokens.

\paragraph{Semantic-guided feature aggregation.}
Learnable queries $\mathbf Q^{(0)}\in\mathbb R^{4\times d}$, shared across instances, represent coordinate roles $\mathcal C=\{x_0,y_0,x_1,y_1\}$, with row $c$ denoted by $\mathbf q_c^{(0)}$. The subscript $i$ indexes instances, while superscripts $(0)$--$(3)$ indicate successive query-update stages rather than powers. Operators $\operatorname{CA}_s$ and $\operatorname{CA}_v$ denote multi-head cross-attention over semantic and visual features, respectively. The first argument supplies queries, and the second supplies keys and values; each operator includes a residual connection and RMS normalization. Queries first read semantic features through layer cross-attention, then image evidence through image cross-attention, followed by coordinate interaction:
\begin{equation}
\begin{aligned}
\mathbf Q_i^{(1)} &= \operatorname{CA}_s(\mathbf Q^{(0)},\mathbf S_i),\\
\mathbf Q_i^{(2)} &= \operatorname{CA}_v(\mathbf Q_i^{(1)},\mathbf U),\\
\mathbf Q_i^{(3)} &= \operatorname{CoordBlock}(\mathbf Q_i^{(2)}).
\end{aligned}
\label{eq:semantic_visual_attention}
\end{equation}
Semantic aggregation thus conditions subsequent visual retrieval. The lower panel of Fig.~\ref{fig:model_framework} illustrates this sequence for the dog instance ($i=1$).

\paragraph{Coordinate interaction.}
$\operatorname{CoordBlock}$ applies self-attention among the four coordinate queries and a residual feed-forward network (FFN), transforming $\mathbf Q_i^{(2)}$ into $\mathbf Q_i^{(3)}$.

\paragraph{Parallel coordinate classification.}
A shared affine classifier $f_o:\mathbb R^d\rightarrow\mathbb R^K$, separate from the MLLM language head, maps each updated coordinate query to a distribution over $K$ bins:
\begin{equation}
\begin{aligned}
\mathbf p_{i,c} &= \operatorname{softmax}(f_o(\mathbf q_{i,c}^{(3)})),\\
\hat b_{i,c} &= \frac{\operatorname*{arg\,max}_{k\in\{0,\ldots,K-1\}}p_{i,c}(k)}{K-1}.
\end{aligned}
\label{eq:coordinate_classification}
\end{equation}
Here $\mathbf q_{i,c}^{(3)}$ is row $c$ of $\mathbf Q_i^{(3)}$, and $p_{i,c}(k)$ is the probability that coordinate $c$ of instance $i$ occupies bin $k\in\{0,\ldots,K-1\}$, corresponding to normalized position $k/(K-1)$; we use $K=1000$. The four distributions are predicted in parallel, with dependencies modeled by query interaction rather than autoregressive coordinate generation. Normalized boxes are scaled by image width $W$ and height $H$.

The decoder contains 11.02M parameters, excluding the backbone and proposal-token vocabulary extensions, and reuses the hidden states used to predict each proposal token and visual features computed during autoregressive generation.

\subsection{Learning Spatial Proposals and Box Geometry}
\label{sec:joint_learning}

The first SFT stage learns grounding from a multi-task data mixture, and the second rebalances the mixture. GRPO post-training then trains proposal generation and geometric decoding on model-generated outputs.

\paragraph{Supervised learning.}
\label{sec:joint_sft}
SFT supervises instance generation and box geometry: the autoregressive loss $\mathcal L_{\mathrm{AR}}$ supervises answer tokens, while target serialization aligns proposals with ground-truth boxes. A one-hot coordinate target does not explicitly encode how far an incorrect bin lies from the target. We therefore use coordinate classification with Gaussian soft labels to account for distances between coordinate values; related smoothing appears in keypoint localization~\cite{simcc}. For normalized ground-truth coordinate $b^*_{i,c}$, let $\mu_{i,c}=(K-1)b^*_{i,c}$ and define the target probability for bin $k$ as
\begin{equation}
p^*_{i,c}(k)=
\frac{\exp[-(k-\mu_{i,c})^2/(2\sigma^2)]}
{\sum_{j=0}^{K-1}\exp[-(j-\mu_{i,c})^2/(2\sigma^2)]}.
\label{eq:gaussian_target}
\end{equation}
Here $p^*_{i,c}(k)$ is normalized over the $K$ bins; the center $\mu_{i,c}$ is continuous without rounding, and $\sigma$ is measured in bins. With coordinate cross-entropy (CE) $\mathcal L_{\mathrm{coord}}$ averaged over target instances and coordinates, the SFT objective is
\begin{equation}
\mathcal L_{\mathrm{SFT}}=\mathcal L_{\mathrm{AR}}
+\lambda_{\mathrm{coord}}\mathcal L_{\mathrm{coord}}.
\label{eq:sft_loss}
\end{equation}
Here, $\lambda_{\mathrm{coord}}=1.0$. Coordinate supervision uses only cross-entropy and updates both the decoder and the MLLM. Proposal prediction and coordinate classification shape the representations used for box decoding, coupling coarse spatial supervision with fine-grained localization. Empty queries retain autoregressive rejection supervision but contribute no coordinate loss.

\begin{table*}[!t]
\centering
\caption{{Common and long-tailed object detection on LVIS and COCO.}}
\label{tab:common_detection}
\begingroup
\footnotesize
\setlength{\tabcolsep}{5pt}
\renewcommand{\arraystretch}{1.08}
\setlength{\aboverulesep}{0pt}
\setlength{\belowrulesep}{0pt}
\scalebox{0.845}{%
\begin{tabular}{@{}c|c|c|c|ccc|c|ccc@{}}
\toprule
\multicolumn{1}{@{}c|}{\multirow[c]{3}{*}{\textbf{Type}}} & \multirow[c]{3}{*}{\textbf{Method}} & \multirow[c]{3}{*}{\shortstack{\textbf{Score}\\\textbf{Thresh.}}} & \multicolumn{4}{c|}{\textbf{LVIS}} & \multicolumn{4}{c}{\textbf{COCO}} \\
\cline{4-7} \cline{8-11}
  & & & \shortstack{\rule{0pt}{\ht\strutbox}Zero-Shot\\(LVIS)} & \shortstack{\rule{0pt}{\ht\strutbox}F1@Iou\\0.5} & \shortstack{\rule{0pt}{\ht\strutbox}F1@Iou\\0.95} & \shortstack{\rule{0pt}{\ht\strutbox}F1@Iou\\mIou} & \shortstack{\rule{0pt}{\ht\strutbox}Zero-Shot\\(COCO)} & \shortstack{\rule{0pt}{\ht\strutbox}F1@Iou\\0.5} & \shortstack{\rule{0pt}{\ht\strutbox}F1@Iou\\0.95} & \shortstack{\rule{0pt}{\ht\strutbox}F1@Iou\\mIou} \\
\midrule
Open-set & Grounding DINO-Swin-T~\cite{grounding_dino} & 0.21 / 0.37 & Yes & 47.70 & 22.70 & 38.80 & Yes & 69.80 & \underline{23.00} & \underline{56.60} \\
\midrule
\multirow[c]{5}{*}{Closed-set} & Faster RCNN-R50~\cite{faster_rcnn} & 0.42 & -- & -- & -- & -- & No & 60.60 & 7.10 & 48.10 \\
 & DETR-R50~\cite{detr} & 0.78 & -- & -- & -- & -- & No & 65.90 & 13.60 & 48.30 \\
 & Deformable-DETR-R50~\cite{deformable_detr} & 0.34 & -- & -- & -- & -- & No & 69.70 & 17.70 & 54.70 \\
 & DINO-R50~\cite{dino} & 0.30 & -- & -- & -- & -- & No & 68.80 & 21.10 & 55.60 \\
 & DINO-Swin-L~\cite{dino} & 0.32 & -- & -- & -- & -- & No & \textbf{75.60} & \textbf{25.40} & \textbf{62.10} \\
\midrule
\multirow[c]{11}{*}{MLLM} & DeepSeek-VL2-Small~\cite{deepseek_vl2} & -- & -- & 56.20 & 21.00 & 41.80 & -- & 60.90 & 14.90 & 45.90 \\
 & MiMo-VL-7B~\cite{mimo_vl} & -- & -- & 49.50 & 8.80 & 31.40 & -- & 56.50 & 6.70 & 35.90 \\
 & OVIS2.5-2B~\cite{ovis_2_5} & -- & -- & 54.40 & 15.80 & 37.40 & -- & 56.20 & 10.30 & 38.70 \\
 & Qwen3-VL-4B~\cite{qwen3_vl} & -- & -- & 59.80 & 20.00 & 43.50 & -- & 63.00 & 14.20 & 46.10 \\
 & Qwen3-VL-8B~\cite{qwen3_vl} & -- & -- & 61.50 & 20.20 & 44.80 & -- & 62.80 & 14.00 & 45.70 \\
 & Cosmos-Reason2-8B~\cite{cosmos_reason2} & -- & -- & 56.40 & 9.80 & 40.20 & -- & 56.40 & 9.80 & 39.30 \\
 & SEED1.5-VL~\cite{seed_1_5_vl} & -- & Yes & 65.60 & 19.50 & 46.70 & Yes & 71.30 & 14.30 & 51.40 \\
 & Rex-Omni-3B~\cite{rex_omni} & -- & Yes & 64.30 & 20.70 & 46.90 & Yes & 72.00 & 15.90 & 52.90 \\
 & LocateAnything-3B~\cite{locate_anything} & -- & Yes & 62.30 & \textbf{31.10} & 50.70 & Yes & 70.10 & 19.30 & 54.70 \\
\cline{2-11}
 & \textbf{SemGeo-SFT} & -- & Yes & \underline{68.15} & 29.31 & \underline{53.23} & Yes & 69.99 & 19.65 & 54.12 \\
 & \textbf{SemGeo} & -- & Yes & \textbf{71.30} & \underline{29.93} & \textbf{55.32} & Yes & \underline{73.27} & 19.95 & 56.42 \\
\bottomrule
\end{tabular}%
}
\endgroup
\end{table*}

\begin{table*}[!t]
\centering
\caption{{Dense object detection and referring object detection.}}
\label{tab:dense_referring_grounding}
\begingroup
\footnotesize
\setlength{\tabcolsep}{1.7pt}
\renewcommand{\arraystretch}{1.08}
\setlength{\aboverulesep}{0pt}
\setlength{\belowrulesep}{0pt}
\resizebox{0.845\textwidth}{!}{%
\begin{tabular*}{\textwidth}{@{\extracolsep{\fill}}c|c|ccc|ccc|ccc|ccc@{}}
\toprule
\multicolumn{1}{@{}c|}{\multirow[c]{4}{*}{\textbf{Type}}} & \multirow[c]{4}{*}{\textbf{Method}} & \multicolumn{6}{c|}{\textbf{Dense Object Detection}} & \multicolumn{6}{c}{\textbf{Referring Object Detection}} \\
\cline{3-8} \cline{9-14}
 & & \multicolumn{3}{c|}{\textbf{Dense200}} & \multicolumn{3}{c|}{\textbf{VisDrone}} & \multicolumn{3}{c|}{\textbf{HumanRef}} & \multicolumn{3}{c}{\textbf{RefCOCOg test}} \\
\cline{3-5} \cline{6-8} \cline{9-11} \cline{12-14}
 & & \shortstack{\rule{0pt}{\ht\strutbox}F1@Iou\\0.5} & \shortstack{\rule{0pt}{\ht\strutbox}F1@Iou\\0.95} & \shortstack{\rule{0pt}{\ht\strutbox}F1@Iou\\mIou} & \shortstack{\rule{0pt}{\ht\strutbox}F1@Iou\\0.5} & \shortstack{\rule{0pt}{\ht\strutbox}F1@Iou\\0.95} & \shortstack{\rule{0pt}{\ht\strutbox}F1@Iou\\mIou} & \shortstack{\rule{0pt}{\ht\strutbox}F1@Iou\\0.5} & \shortstack{\rule{0pt}{\ht\strutbox}F1@Iou\\0.95} & \shortstack{\rule{0pt}{\ht\strutbox}F1@Iou\\mIou} & \shortstack{\rule{0pt}{\ht\strutbox}F1@Iou\\0.5} & \shortstack{\rule{0pt}{\ht\strutbox}F1@Iou\\0.95} & \shortstack{\rule{0pt}{\ht\strutbox}F1@Iou\\mIou} \\
\midrule
Open-set & Grounding DINO-Swin-T~\cite{grounding_dino} & 36.90 & \underline{19.70} & 33.10 & 55.20 & \textbf{3.90} & 38.50 & 28.00 & 16.50 & 25.20 & 53.80 & 22.90 & 46.80  \\
\midrule
\multirow[c]{11}{*}{MLLM} & OVIS2.5-2B~\cite{ovis_2_5} & 17.90 & 0.00 & 6.70 & 21.00 & 0.10 & 9.20 & 70.60 & 12.30 & 50.00 & 87.60 & 30.50 & 73.80  \\
 & MiMo-VL-7B~\cite{mimo_vl} & 29.70 & 0.40 & 15.90 & 27.70 & 0.30 & 14.30 & 77.60 & 26.40 & 63.40 & 84.60 & 14.90 & 65.50  \\
 & DeepSeek-VL2-Small~\cite{deepseek_vl2} & 16.00 & 3.90 & 12.70 & 35.80 & 1.70 & 23.30 & 72.00 & 46.50 & 64.70 & \textbf{91.80} & \textbf{47.00} & \textbf{81.60}  \\
 & Qwen3-VL-4B~\cite{qwen3_vl} & 17.50 & 2.40 & 12.50 & 42.30 & 1.40 & 26.00 & 77.70 & 54.90 & 71.10 & 87.60 & 33.90 & 74.60  \\
 & Qwen3-VL-8B~\cite{qwen3_vl} & 13.50 & 1.70 & 9.60 & 42.80 & 1.40 & 25.80 & 78.60 & 55.70 & 72.00 & 88.60 & 33.80 & 75.20  \\
 & Cosmos-Reason2-8B~\cite{cosmos_reason2} & 25.10 & 1.10 & 15.10 & 40.20 & 1.30 & 22.30 & -- & -- & -- & -- & -- & -- \\
 & SEED1.5-VL~\cite{seed_1_5_vl} & \underline{76.90} & 5.30 & 53.20 & 55.90 & 0.60 & 27.40 & \textbf{88.20} & 60.00 & \textbf{81.60} & 85.20 & 32.10 & 73.20  \\
 & Rex-Omni-3B~\cite{rex_omni} & \textbf{78.40} & 10.30 & 58.30 & 61.60 & 1.50 & 35.80 & \underline{85.40} & \underline{65.40} & \underline{79.90} & 86.80 & 36.60 & 74.30  \\
 & LocateAnything-3B~\cite{locate_anything} & 74.00 & 18.50 & \underline{58.70} & \underline{63.00} & \underline{3.20} & \underline{39.90} & 82.90 & \textbf{68.80} & 78.70 & \underline{88.80} & \underline{43.40} & \underline{77.60}  \\
\cline{2-14}
 & \textbf{SemGeo-SFT} & 57.00 & 15.00 & 45.40 & 50.50 & 2.40 & 31.90 & 80.90 & 62.60 & 75.00 & 86.90 & 40.10 & 74.60  \\
 & \textbf{SemGeo} & \textbf{78.40} & \textbf{19.90} & \textbf{63.40} & \textbf{65.60} & 3.00 & \textbf{41.60} & 82.50 & 64.80 & 77.20 & 88.30 & 41.90 & 76.30  \\
\bottomrule
\end{tabular*}%
}
\endgroup
\end{table*}

\paragraph{Reinforcement post-training.}
\label{sec:joint_post_training}
Gaussian soft labels account for distances between coordinate values, but do not directly optimize the quality of the final predicted box set. Moreover, teacher-forced training conditions instance generation and geometric decoding on ground-truth prefixes, whereas inference relies on previously generated tokens. This \emph{exposure bias}~\cite{mixer} affects both the generation policy and the hidden states consumed by the decoder. We therefore combine GRPO~\cite{deepseek_math}, which provides rewards for the predicted box set, with coordinate supervision on generated proposals matched to ground-truth boxes. Coordinate CE with Gaussian soft labels adapts the decoder and its supporting MLLM representations to model-generated outputs.

For a valid response with $N$ predictions and $N^*>0$ ground-truth targets, we solve two optimal bipartite matching problems within each category using linear sum assignment. The F1 matching maximizes the number of pairs satisfying IoU $\geq\tau$, yielding $\mathrm{TP}_{\tau}$. Separately, $\mathcal M$ maximizes total IoU at maximum cardinality without an IoU threshold. The three rewards measure correctness of the predicted box set, box localization quality, and coarse proposal alignment:
\begin{equation}
\begin{aligned}
R_{\mathrm{F1}} &= {2\mathrm{TP}_{\tau}}/{(N+N^*)},\\
R_{\mathrm{box}} &= {2\sum_{(i,j)\in\mathcal M}u_{ij}}/{(N+N^*)},\\
R_{\mathrm{grid}} &= {2\sum_{(i,j)\in\mathcal M}a_{ij}}/{(N+N^*)}.
\end{aligned}
\label{eq:reward_components}
\end{equation}
Here $u_{ij}$ is box IoU, and $a_{ij}$ is a binary grid-hit indicator that equals one when the proposal cell overlaps its assigned target box with positive area and zero otherwise. Both are zero for invalid predicted boxes, which still count toward $N$. Unmatched predictions also remain in the denominator. One-to-one matching prevents multiple predictions from receiving credit for the same target, while normalization by $N+N^*$ discourages redundant predictions and missed targets. Continuous IoU provides graded localization feedback beyond the thresholded F1 signal, and the grid term encourages spatial alignment between proposals and matched targets. The total outcome reward is
\begin{equation}
R=w_fR_{\mathrm{F1}}+w_bR_{\mathrm{box}}+w_gR_{\mathrm{grid}},
\quad w_f+w_b+w_g=1.
\label{eq:outcome_reward}
\end{equation}
We use $\tau=0.5$ and $(w_f,w_b,w_g)=(0.6,0.3,0.1)$.

Group-normalized rewards provide advantages for the clipped policy objective. We reuse the same-category one-to-one assignment $\mathcal M$, without an IoU threshold, to supervise matched proposals with coordinate CE using Gaussian soft labels. Low-IoU and geometrically invalid predictions therefore receive coordinate correction when assigned a target, whereas unmatched proposals receive no coordinate supervision. The post-training objective is
\begin{equation}
\begin{aligned}
\mathcal L_{\mathrm{post}}(\theta,\phi)
&=\mathcal L_{\mathrm{GRPO}}(\theta)+\beta\mathcal L_{\mathrm{KL}}(\theta)\\
&\quad+\lambda_{\mathrm{post}}\mathcal L_{\mathrm{coord}}^{\mathrm{matched}}(\theta,\phi).
\end{aligned}
\label{eq:post_training_loss}
\end{equation}
Here, $\beta=0.01$, $\lambda_{\mathrm{post}}=1.0$, and $\mathcal L_{\mathrm{coord}}^{\mathrm{matched}}$ averages coordinate CE with Gaussian soft labels over matched proposals. Rewards and assignments are fixed from the rollout; coordinate distributions are recomputed using the current MLLM and decoder. The coordinate loss backpropagates through the decoder and its semantic and visual inputs without feature detachment, updating both $\theta$ and $\phi$; policy/KL terms update only $\theta$. Thus, both components learn from model-generated outputs.

\section{Experiments}
\label{sec:experiments}

\subsection{Training Details and Evaluation Setup}
\label{sec:experimental_setup}

\paragraph{Training details.}
We build SemGeo on Qwen3-VL-2B-Instruct~\cite{qwen3_vl} through two SFT stages followed by GRPO post-training. The three stages use configured mixtures of 10.40M, 4.47M, and 100.13K annotation records, with checkpoints selected at 60k, 15k, and 1.3k steps, respectively. Both SFT stages use 24 GPUs, while GRPO post-training uses 8 GPUs. To isolate architectural effects from large-scale multi-task training, all grid and decoder ablations are conducted under an identical, single-dataset~\cite{coco} training protocol. Implementation, training, and evaluation details are provided in the supplementary material.

\paragraph{Compared methods.}
We compare against three groups: (1) open-set detectors, including Grounding DINO~\cite{grounding_dino}; (2) closed-set detectors, including DETR variants~\cite{detr,deformable_detr,dino} and DocLayout-YOLO~\cite{doclayout_yolo}; and (3) MLLMs, including Qwen3-VL~\cite{qwen3_vl}, DeepSeek-VL2~\cite{deepseek_vl2}, OVIS2.5~\cite{ovis_2_5}, MiMo-VL~\cite{mimo_vl}, SEED1.5-VL~\cite{seed_1_5_vl}, Rex-Omni~\cite{rex_omni}, and LocateAnything~\cite{locate_anything}. To compare inference speed, we re-evaluate Qwen3-VL-2B, Rex-Omni, and LocateAnything under the listed backends and measure accuracy together with timing. We report SemGeo-SFT after the second SFT stage and SemGeo after GRPO post-training, both using native vLLM inference in the main accuracy tables.

\paragraph{Evaluation protocol.}
Following the evaluation framework of Rex-Omni~\cite{rex_omni} and LocateAnything ~\cite{locate_anything}, we evaluate object detection on COCO~\cite{coco}, LVIS~\cite{lvis}, Dense200, and VisDrone~\cite{visdrone}; referring object detection on HumanRef~\cite{human_ref} and RefCOCOg test~\cite{refcocog}; document layout grounding on DocLayNet~\cite{doclaynet} and M6Doc~\cite{m6doc}. SemGeo-SFT and SemGeo are evaluated zero-shot on COCO and LVIS. We report F1@0.5, F1@0.95, and threshold-aggregated F1@mIou over IoU thresholds 0.5:0.05:0.95, without requiring calibrated confidence scores. Scores and absolute differences are expressed in percent. In the main accuracy tables, \textbf{bold} and \underline{underlined} values mark the best and second-best results across methods, with ties sharing a rank; ``--'' denotes unknown, unreported, or unavailable entries.

\subsection{Experimental Results}
\label{sec:main_results}

\paragraph{Common and long-tailed detection.}
SemGeo achieves 56.42\% F1@mIou on COCO and 55.32\% on LVIS, surpassing LocateAnything by 1.72\% and 4.62\% and Rex-Omni by 3.52\% and 8.42\%, respectively (Table~\ref{tab:common_detection}). These are the highest F1@mIou scores among the compared MLLMs on both datasets. On COCO, SemGeo approaches Grounding DINO's 56.60\% F1@mIou, while the closed-set DINO-Swin-L remains higher at 62.10\%. On LVIS, LocateAnything retains a higher F1@0.95 (31.10\% vs.\ 29.93\%).

\paragraph{Dense object detection.}
SemGeo obtains 63.40\% and 41.60\% F1@mIou on Dense200 and VisDrone, respectively, outperforming LocateAnything by 4.70\% and 1.70\% and Rex-Omni by 5.10\% and 5.80\% (Table~\ref{tab:dense_referring_grounding}). On Dense200, SemGeo matches Rex-Omni's F1@0.5 (78.40\%) and achieves the highest reported F1@0.95 (19.90\%). Grounding DINO remains higher at IoU 0.95 on VisDrone (3.90\% vs.\ 3.00\%). Overall, SemGeo achieves the strongest F1@mIou among the compared methods in these dense and small-object scenarios.

\begin{table}[!t]
\centering
\caption{{Document layout grounding on DocLayNet and M6Doc.}}
\label{tab:layout_grounding_compact}
\begingroup
\footnotesize
\setlength{\tabcolsep}{1.2pt}
\renewcommand{\arraystretch}{1.08}
\setlength{\aboverulesep}{0pt}
\setlength{\belowrulesep}{0pt}
\resizebox{0.875\columnwidth}{!}{%
\begin{tabular*}{\columnwidth}{@{\extracolsep{\fill}}c|c|cc|cc@{}}
\toprule
\multicolumn{1}{@{}c|}{\multirow[c]{3}{*}{\textbf{Type}}} & \multirow[c]{3}{*}{\textbf{Method}} & \multicolumn{2}{c|}{\textbf{DocLayNet}} & \multicolumn{2}{c}{\textbf{M6Doc}} \\
\cline{3-4} \cline{5-6}
 & & {\scriptsize\shortstack{\rule{0pt}{\ht\strutbox}F1@Iou\\0.5}} & {\scriptsize\shortstack{\rule{0pt}{\ht\strutbox}F1@Iou\\mIou}} & {\scriptsize\shortstack{\rule{0pt}{\ht\strutbox}F1@Iou\\0.5}} & {\scriptsize\shortstack{\rule{0pt}{\ht\strutbox}F1@Iou\\mIou}} \\
\midrule
Closed-set & DocLayout-YOLO~\cite{doclayout_yolo} & \underline{91.20} & \textbf{81.10} & -- & -- \\
\midrule
\multirow[c]{7}{*}{MLLM} & Qwen3-VL-4B~\cite{qwen3_vl} & 60.80 & 37.20 & 30.60 & 19.00 \\
 & Qwen3-VL-8B~\cite{qwen3_vl} & 54.70 & 34.10 & 37.20 & 22.70 \\
 & SEED1.5-VL~\cite{seed_1_5_vl} & 54.90 & 28.70 & 48.00 & 28.00 \\
 & Rex-Omni-3B~\cite{rex_omni} & 89.50 & 70.70 & 76.30 & 55.60 \\
 & LocateAnything-3B~\cite{locate_anything} & 91.10 & 76.80 & \textbf{90.60} & \underline{70.10} \\
\cline{2-6}
 & \textbf{SemGeo-SFT} & 90.40 & 76.90 & \underline{81.80} & 65.00 \\
 & \textbf{SemGeo} & \textbf{94.30} & \underline{79.70} & \textbf{90.60} & \textbf{73.20} \\
\bottomrule
\end{tabular*}%
}
\endgroup
\end{table}

\paragraph{Referring object detection.}
SemGeo reaches 77.20\% and 76.30\% F1@mIou on HumanRef and RefCOCOg test (Table~\ref{tab:dense_referring_grounding}). On RefCOCOg test, it improves over Rex-Omni by 2.00\%, with F1@0.95 increasing from 36.60\% to 41.90\%. Compared with LocateAnything, F1@mIou is lower by 1.50\% and 1.30\% on HumanRef and RefCOCOg test, respectively. 
Further qualitative inspection reveals that target misidentification in some named-entity queries (e.g., people's names). The larger gap at F1@0.95 than at F1@0.5 is consistent with observed boxes extending beyond annotated regions for some correctly identified targets, resulting in lower IoU. 

\paragraph{Document layout grounding.}
SemGeo achieves 79.70\% and 73.20\% F1@mIou on DocLayNet and M6Doc, exceeding LocateAnything by 2.90\% and 3.10\%, respectively (Table~\ref{tab:layout_grounding_compact}). These results demonstrate effective structured document localization beyond natural-image detection. The specialized DocLayout-YOLO retains a higher F1@mIou on DocLayNet, while SemGeo achieves a higher F1@0.5 (94.30\% vs.\ 91.20\%).

\begin{figure}[!t]
\centering
\includegraphics[width=0.85\columnwidth]{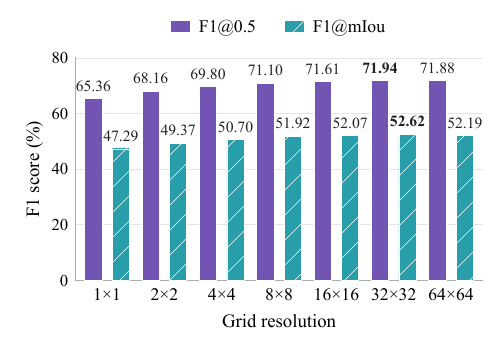}
\caption{\textbf{Effect of spatial proposal granularity on COCO.} The $1\times1$ setting uses a single shared proposal token.}
\label{fig:grid_ablation}
\end{figure}

\begin{table}[!t]
\centering
\caption{\textbf{Geometric decoder ablations on COCO.} Full combines four language layers, semantic and visual cross-attention, coordinate self-attention, and a shared coordinate classifier. Coordinate targets are Gaussian soft labels or one-hot labels. }
\label{tab:decoder_ablation}
\begingroup
\small
\setlength{\tabcolsep}{3pt}
\renewcommand{\arraystretch}{1.06}
\setlength{\aboverulesep}{0pt}
\setlength{\belowrulesep}{0pt}
\resizebox{0.85\columnwidth}{!}{%
\begin{tabular}{@{}c|c|c|cc@{}}
\toprule
\multicolumn{1}{@{}c|}{\makebox[\dimexpr0.40\columnwidth-\tabcolsep-\arrayrulewidth\relax][c]{\raisebox{0.4\baselineskip}[0pt][0pt]{\textbf{Decoder variant}}}} & \makebox[\dimexpr0.19\columnwidth-2\tabcolsep-\arrayrulewidth\relax][c]{\shortstack{\rule{0pt}{\ht\strutbox}Coordinate\\targets}} & \makebox[\dimexpr0.13\columnwidth-2\tabcolsep-\arrayrulewidth\relax][c]{\shortstack{\rule{0pt}{\ht\strutbox}Params\\(M)}} & \makebox[\dimexpr0.14\columnwidth-2\tabcolsep\relax][c]{\shortstack{\rule{0pt}{\ht\strutbox}F1@Iou\\0.5}} & \makebox[\dimexpr0.14\columnwidth-\tabcolsep\relax][c]{\shortstack{\rule{0pt}{\ht\strutbox}F1@Iou\\mIou}} \\
\midrule
Last-layer MLP-2048 & Gaussian & 41.76 & 70.73 & 50.43 \\
Last-layer MLP-512 & Gaussian & 5.20 & 66.22 & 45.37 \\
\midrule
Last layer + visual & Gaussian & 7.87 & 67.95 & 47.19 \\
Four layers, no visual & Gaussian & 8.92 & 70.03 & 50.28 \\
\midrule
Four layers + visual (Full) & One-hot & 11.02 & 71.45 & 51.88 \\
Four layers + visual (Full) & Gaussian & 11.02 & \textbf{71.94} & \textbf{52.62} \\
\bottomrule
\end{tabular}%
}
\endgroup
\end{table}

\begin{table*}[!t]
\centering
\caption{\textbf{Accuracy and inference speed comparison on COCO.} Baselines are re-evaluated locally under the listed backend. All metrics are three-run means, with sample standard deviations for timing and mean BPS. LocateAnything uses \texttt{la\_flash} in Hybrid mode. Box counts are not ranked. }
\label{tab:efficiency_comparison}
\begingroup
\footnotesize
\setlength{\tabcolsep}{2pt}
\renewcommand{\arraystretch}{1.05}
\setlength{\aboverulesep}{0pt}
\setlength{\belowrulesep}{0pt}
\resizebox{0.85\textwidth}{!}{%
\begin{tabular}{@{}c|c|c|cc|cc|ccc@{}}
\toprule
\textbf{Backend} & \textbf{Method} & \textbf{Size} & \shortstack{\rule{0pt}{\ht\strutbox}Recall@Iou\\0.5} & \shortstack{\rule{0pt}{\ht\strutbox}F1@Iou\\0.5} & \shortstack{\rule{0pt}{\ht\strutbox}Mean Tokens\\per Image$\downarrow$} & \shortstack{\rule{0pt}{\ht\strutbox}Mean Boxes\\per Image} & \shortstack{\rule{0pt}{\ht\strutbox}Generation time\\Mean (ms)$\downarrow$} & \shortstack{\rule{0pt}{\ht\strutbox}Generation time\\P95 (ms)$\downarrow$} & \shortstack{\rule{0pt}{\ht\strutbox}Mean BPS\\(boxes/s)$\uparrow$} \\
\midrule
\multirow[c]{5}{*}{Transformers} & Qwen3-VL~\cite{qwen3_vl} & 2B & 54.19 & 63.87 & $243.84$ & $4.74$ & $7826.5\pm168.1$ & $22222.8\pm572.0$ & $0.73\pm0.02$ \\
 & Rex-Omni-Style & 2B & \underline{69.07} & 70.29 & $57.05$ & $7.72$ & $1701.1\pm15.1$ & $4593.5\pm50.7$ & $3.95\pm0.02$ \\
 & Rex-Omni~\cite{rex_omni} & 3B & 68.12 & \underline{71.52} & $59.89$ & $8.30$ & $2133.9\pm23.3$ & $5287.7\pm46.1$ & $3.23\pm0.04$ \\
 & LocateAnything~\cite{locate_anything} & 3B & 64.77 & 70.81 & $\underline{52.34}$ & $6.88$ & $\underline{964.7\pm3.6}$ & $\underline{2565.6\pm11.1}$ & $\underline{6.08\pm0.02}$ \\
 & \textbf{SemGeo} & 2B & \textbf{70.37} & \textbf{73.24} & $\mathbf{29.34}$ & $8.05$ & $\mathbf{837.6\pm9.4}$ & $\mathbf{1895.3\pm20.3}$ & $\mathbf{7.62\pm0.07}$ \\
\midrule
\multirow[c]{5}{*}{vLLM} & Qwen3-VL~\cite{qwen3_vl} & 2B & 53.81 & 64.17 & $239.81$ & $4.54$ & $1532.4\pm5.0$ & $5233.3\pm6.5$ & $3.87\pm0.01$ \\
 & Rex-Omni-Style & 2B & \underline{68.72} & 70.62 & $\underline{56.64}$ & $7.64$ & $\underline{299.6\pm0.3}$ & $\underline{752.7\pm1.9}$ & $\underline{21.25\pm0.02}$ \\
 & Rex-Omni~\cite{rex_omni} & 3B & 68.04 & \underline{71.48} & $59.84$ & $8.29$ & $485.2\pm1.4$ & $1137.1\pm2.1$ & $13.41\pm0.05$ \\
 & LocateAnything~\cite{locate_anything} & 3B & -- & -- & -- & -- & -- & -- & -- \\
 & \textbf{SemGeo} & 2B & \textbf{70.40} & \textbf{73.27} & $\mathbf{29.76}$ & $8.46$ & $\mathbf{223.6\pm1.3}$ & $\mathbf{411.1\pm0.9}$ & $\mathbf{27.74\pm0.16}$ \\
\bottomrule
\end{tabular}%
}
\endgroup
\end{table*}

\subsection{Ablation Studies}
\label{sec:ablations}

\paragraph{Spatial proposal granularity.}
We vary the grid resolution from $1\times1$ to $64\times64$ to evaluate the effect of coarse spatial supervision on grounding (Fig.~\ref{fig:grid_ablation}). A $1\times1$ grid assigns every instance the same proposal token, removing cell-level spatial distinctions while retaining geometric decoding. It achieves 65.36\% F1@0.5 and 47.29\% F1@mIou, compared with 71.94\% and 52.62\% for $32\times32$. With one proposal token per instance in both variants, the 5.33\% F1@mIou gain supports spatially differentiated proposals for coarse-to-fine grounding. Increasing the resolution to $64\times64$ yields no further improvement, with F1@mIou decreasing to 52.19\%, while expanding the grid-token vocabulary from 1024 to 4096 entries. We therefore choose $32\times32$, which achieves the highest F1 scores with a smaller vocabulary than $64\times64$.

\paragraph{Geometric decoding and coordinate supervision.}
To evaluate the decoder design and coordinate supervision, we compare MLP heads, semantic and visual feature combinations, and coordinate target distributions (Table~\ref{tab:decoder_ablation}). The full decoder exceeds the single-layer visual-attention variant and the multi-layer variant without visual attention by 5.43\% and 2.34\% in F1@mIou, respectively. It also outperforms MLP-2048 by 2.19\% with 73.6\% fewer decoder parameters, supporting structured feature integration rather than simply increasing capacity. With the full architecture fixed, Gaussian soft labels improve F1@0.5 and F1@mIou over one-hot targets by 0.49\% and 0.74\%, respectively. Together, these results support combining multi-layer semantic and visual features with coordinate supervision that accounts for distances between coordinate values.

\subsection{Accuracy and Inference Speed}
\label{sec:efficiency}

Table~\ref{tab:efficiency_comparison} compares SemGeo with Qwen3-VL-2B, Rex-Omni~(3B), and LocateAnything~(3B). To compare methods built on the same backbone, we also fine-tune Qwen3-VL-2B on COCO using Rex-Omni's coordinate-token formulation, denoted Rex-Omni-Style~(2B). The other baselines are evaluated using their existing checkpoints without additional fine-tuning. We use one GPU, batch size 1, and a shared hardware and software environment, reporting Transformers~\cite{transformers} and vLLM~\cite{vllm} separately. Following LocateAnything~\cite{locate_anything}, we measure inference speed using mean BPS, the mean per-image box-generation rate. We also report mean and P95 generation times, including geometric decoding, with paired recall and F1.

Under vLLM, SemGeo reduces output tokens from 239.81 to 29.76 per image and mean generation time from 1532.4 to 223.6\,ms relative to Qwen3-VL-2B, while improving recall and F1@0.5 by 16.59\% and 9.10\%, respectively. Reusing features computed during generation without an additional backbone pass, SemGeo reaches 27.74 mean BPS (2.07$\times$ Rex-Omni and 1.31$\times$ Rex-Omni-Style) and generates 47.5\% fewer tokens than the latter. Under Transformers, it outperforms LocateAnything by 25.4\% in mean BPS and 2.43\% in F1@0.5, while reducing mean generation time and P95 by 13.2\% and 26.1\%, respectively. SemGeo combines higher accuracy with lower generation latency and higher box-generation throughput.

\begin{figure}[!t]
\centering
\begingroup
\setlength{\fboxsep}{0pt}
\setlength{\fboxrule}{0.3pt}
\resizebox{0.85\columnwidth}{!}{%
\begin{minipage}{\columnwidth}
\centering
\begin{minipage}[t]{0.48\linewidth}
\centering
{\small\textbf{SemGeo-SFT}}\par\smallskip
\fbox{\includegraphics[width=\dimexpr\linewidth-2\fboxrule\relax]{figures/example_a_sft.png}}
\par\vspace{2pt}
{\footnotesize Pred: 3}
\end{minipage}\hfill
\begin{minipage}[t]{0.48\linewidth}
\centering
{\small\textbf{SemGeo}}\par\smallskip
\fbox{\includegraphics[width=\dimexpr\linewidth-2\fboxrule\relax]{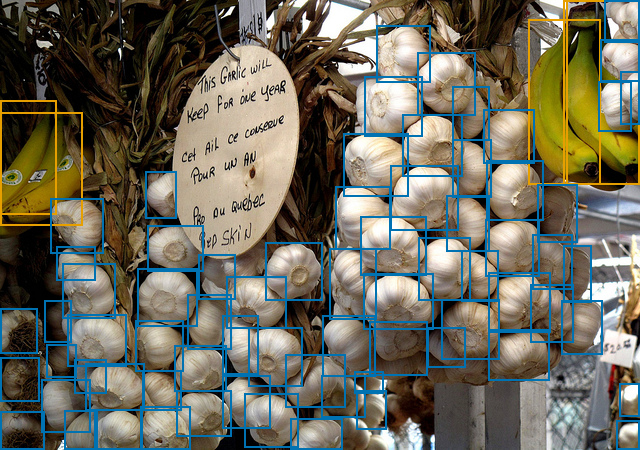}}
\par\vspace{2pt}
{\footnotesize Pred: 76}
\end{minipage}
\par\vspace{3pt}
{\footnotesize (a) Query: \textcolor[HTML]{E69F00}{banana}, \textcolor[HTML]{0072B2}{garlic}}
\par\vspace{6pt}
\begin{minipage}[t]{0.48\linewidth}
\centering
\fbox{\includegraphics[width=\dimexpr\linewidth-2\fboxrule\relax]{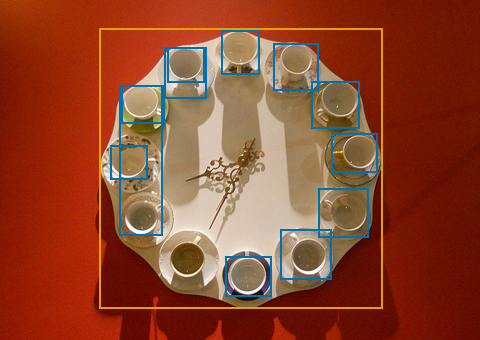}}
\par\vspace{2pt}
{\footnotesize Pred: 13}
\end{minipage}\hfill
\begin{minipage}[t]{0.48\linewidth}
\centering
\fbox{\includegraphics[width=\dimexpr\linewidth-2\fboxrule\relax]{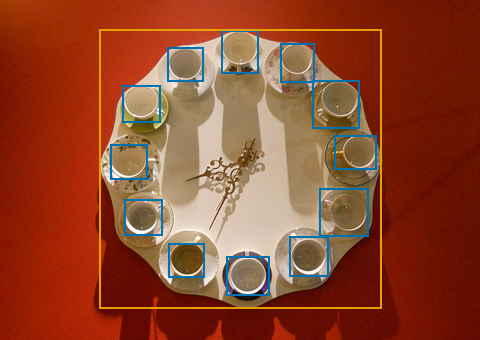}}
\par\vspace{2pt}
{\footnotesize Pred: 13}
\end{minipage}
\par\vspace{3pt}
{\footnotesize (b) Query: \textcolor[HTML]{E69F00}{clock}, \textcolor[HTML]{0072B2}{teacup}}
\end{minipage}%
}%
\endgroup
\caption{\textbf{Qualitative comparison of SemGeo-SFT and SemGeo} on two LVIS examples. All geometrically valid predictions are drawn without deduplication; exact repetitions overlap. Colors follow the queries. Pred counts all outputs, including invalid boxes; GT counts queried instances.}
\label{fig:qualitative_comparison}
\end{figure}

\subsection{Reinforcement Post-training Analysis}
\label{sec:qualitative}

Reinforcement post-training provides feedback on missed and redundant predictions through rewards for the predicted box set, while adapting geometric decoding through coordinate supervision on matched proposals. Relative to SemGeo-SFT, it improves all reported F1 scores across the eight benchmarks (Tables~\ref{tab:common_detection}--\ref{tab:layout_grounding_compact}). F1@mIou gains range from 1.70\% to 18.00\%, with the largest improvements on Dense200, VisDrone, and M6Doc.

Fig.~\ref{fig:qualitative_comparison} illustrates improved instance coverage on two LVIS examples. In (a), SemGeo-SFT outputs only three boxes, missing many queried instances; SemGeo provides broader coverage of bananas and garlic. In (b), both models output 13 boxes for 13 annotated instances, but SemGeo-SFT misses lower-left teacups while producing overlapping or elongated boxes for other cups. In contrast, SemGeo covers the missed cups with distinct instance boxes. These examples show that reinforcement post-training can improve instance coverage beyond changes in prediction count.

\section{Conclusion}
\label{sec:conclusion}

We presented SemGeo, a token-efficient grounding framework that connects spatially supervised instance generation with parallel geometric decoding. Grid tokens provide coarse localization, while reused semantic and visual features recover precise coordinates without autoregressive fine-coordinate sequences. Our training pipeline learns proposal generation and box localization together and adapts both components to model-generated outputs. Experiments across eight benchmarks demonstrate strong grounding accuracy, while runtime comparisons show lower generation latency and higher box-generation throughput. These findings show that compact autoregressive outputs can retain explicit spatial supervision and support precise localization. SemGeo offers a practical coarse-to-fine design for multi-instance grounding by separating spatially guided instance generation from fine geometric prediction.

\textbf{Limitations.}
Despite its accuracy and inference-speed advantages, SemGeo does not achieve the best results on referring object detection. Feature dimensionality reduction in the geometric decoder may limit the preservation of fine-grained semantic information. Future work will focus on better preserving this information to improve referring object detection.


\end{document}